\documentclass[a4paper, 10pt, conference]{ieeeconf}      
\usepackage{graphicx}
\usepackage{amssymb}
\usepackage{url}

\usepackage{url}
   \usepackage{array}
\usepackage{times}
\usepackage{epsfig}
\usepackage{graphicx}
\usepackage{amsmath}
\usepackage{amssymb}
\usepackage{array}
\usepackage{multirow}
\usepackage{microtype}
\usepackage{algpseudocode}
\usepackage{bm}
\usepackage{framed}

\algdef{SE}[SUBALG]{Indent}{EndIndent}{}{\algorithmicend\ }%
\algtext*{Indent}
\algtext*{EndIndent} 

\usepackage{FG2017}

\FGfinalcopy 

\title{\LARGE \bf
Pooling Facial Segments to Face: The Shallow and Deep Ends
}

\author{Upal Mahbub$^{*}$\thanks{* First two authors contributed equally} \quad Sayantan Sarkar$^{*}$   \quad Rama Chellappa\\ 
Department of Electrical and Computer Engineering and the Center for Automation Research, \\UMIACS, University of Maryland, College Park, MD 20742\\
\footnotetext{footnote with two references}
{\tt\small \{umahbub, ssarkar2, rama\}@umiacs.umd.edu}}

\IEEEoverridecommandlockouts

\begin{document}

\ifFGfinal
\thispagestyle{empty}
\pagestyle{empty}
\else
\author{Anonymous FG 2017 submission\\-- DO NOT DISTRIBUTE --\\}
\pagestyle{plain}
\fi

\maketitle

\begin{abstract}
Generic face detection algorithms do not perform very well in the mobile domain due to significant presence of occluded and partially visible faces. One promising technique to handle the challenge of partial faces is to design face detectors based on facial segments. In this paper two such face detectors namely, SegFace and DeepSegFace, are proposed that detect the presence of a face given arbitrary combinations of certain face segments. Both methods use proposals from facial segments as input that are found using weak boosted classifiers. SegFace is a shallow and fast algorithm using traditional features, tailored for situations where real time constraints must be satisfied. On the other hand, DeepSegFace is a more powerful algorithm based on a deep convolutional neutral network (DCNN) architecture. DeepSegFace offers certain advantages over other DCNN-based face detectors as it requires relatively little amount of data to train by utilizing a novel data augmentation scheme and is very robust to occlusion by design. Extensive experiments show the superiority of the proposed methods, specially DeepSegFace, over other state-of-the-art face detectors in terms of precision-recall and ROC curve on two mobile face datasets.
\end{abstract}

\section{Introduction}\label{Introduction}
In recent years, there has been substantial progress in the development of efficient and robust face detection techniques mostly because of the remarkable progress in convolutional neural network (CNN) architectures for face detection and the availability of large amount of face data \cite{RRanjan_Hyperface}\cite{CUHK_FD}\cite{YahooMultiview_FD}. Though the general trend of developing new face detectors is centered around detecting faces in unconstrained environments with large variations in pose and illumination \cite{Ramanan:2012:FDP:2354409.2355119}\cite{fddbTech}\cite{LFWTech}\cite{AFLWDataset}, there is also a growing interest for developing face detectors optimized for detecting occluded and partially visible faces from images captured with mobile devices \cite{FSFD_Mahbub}\cite{Sarkar_DeepFeatureFD}. Reliable and fast detection of faces from the front-camera captures of a mobile device is a fundamental step for applications such as active/continuous authentication of the user of a mobile device \cite{VMP_SPM_AA_2016}\cite{AA02_MahbubChellappa_BTAS2016}\cite{umd_Dataset}\cite{Mobio_2012}. 

Although, face-based authentication systems on mobile devices rely heavily on accurate detection of faces prior to verification, most state-of-the-art techniques are ineffective for mobile devices because of the following reasons:
\begin{enumerate}
\item Face images captured by the front camera of the phone are, in many cases, only partially visible \cite{AA02_MahbubChellappa_BTAS2016}. 
Traditional face detectors such as Viola- Jones's \cite{VJFull} and Deformable part model (DPM) \cite{Ramanan:2012:FDP:2354409.2355119} and more advanced face detectors based on deep convolutional neural networks (DCNN) such as Hyperface \cite{RRanjan_Hyperface}, yahoo multiview \cite{YahooMultiview_FD} and CUHK \cite{CUHK_FD} are usually trained on full faces. While they work well on detecting multiple frontal or profile faces of various resolution, they frequently fail to detect partial faces.

\item For active authentication the recall rate needs to be high at very high precision. Many of the available face detectors have a low recall rate even though the precision is high. When operated at high recall, their precision drops rapidly because of excessive false positive detection. 
\item  The algorithm needs to be simple, fast and customizable in order to operate in real-time on a cellular device. While in \cite{Sarkar_DeepFeatureFD} and \cite{AA_Samangouei_CNN} the authors deploy CNNs on mobile GPUs for face detection and verification, most CNN-based detectors, such as \cite{RRanjan_Hyperface} and some generic methods like \cite{Ramanan:2012:FDP:2354409.2355119} are too complex to run on the mobile platform.
\end{enumerate}

\begin{figure}[t]
\centering
\includegraphics[width=0.45\textwidth]{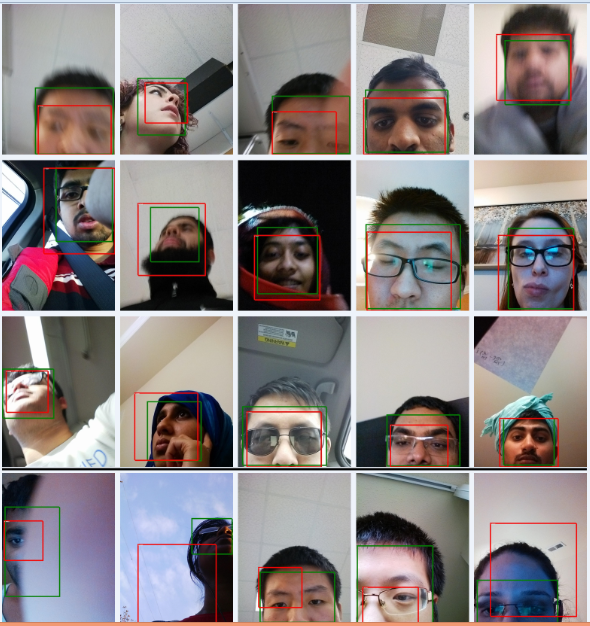}
\caption{Sample face detection outcome of the proposed DeepSegFace method on the UMDAA-02-FD dataset. The first three rows show correct detections at different illumination, pose and partial visibility scenarios, while the last row shows some incorrect detections.}
\vskip -8pt
\label{sampleImages}
\end{figure}

On the other hand, images captured for active authentication offer certain advantages for the face detection problem because of its semi-constrained nature \cite{FSFD_Mahbub}. Usually there is a single user in the frame, hence there is no need to handle multiple face detection. The face is in close proximity of the camera, within a certain range of dimensions and of high resolution, thus eliminating the need for detecting at multiple scales or resolution.

Partial faces such as ones present in images captured by the front camera of mobile devices can be handled if the algorithm is able to effectively combine detections of facial parts into a full face detection. Therefore to address this requirement, two algorithms SegFace and DeepSegFace are proposed in this paper, that detect faces from proposals made of face segments. Sample results for DeepSegFace are shown in Fig. \ref{sampleImages}. The proposal are generated using weak adaboost cascade classifiers following the method described in \cite{FSFD_Mahbub}. 

This paper makes the following contributions:
\begin{enumerate}
\item SegFace, a fast real-time face segment to face detector is proposed.
\item DeepSegFace, a novel deep CNN-based architecture, that detects faces from facial segments-based proposals is developed.
\item A principled scheme is developed that helps augment data as well as regularize the classifier.
\end{enumerate}

In section \ref{RelatedWorks} a summary of works done on face detection in general and in the mobile domain is given. In section \ref{Proposed Method}, the proposed face detection techniques are described in detail. A brief description of the two mobile-face datasets that are used for experimental validation are described in section \ref{Dataset} . All the analysis, experimental results and comparisons for the proposed methods with state-of-the-art methods are provided in section \ref{Experimental Results}. Finally, a brief summary of this work as well as future directions of research are included in section \ref{Conclusion}.

\section{Related Works}\label{RelatedWorks}
Face detection is one of the earliest applications of computer vision dating back several decades \cite{Bledsoe1965}\cite{Kanade1973}. However, most methods before 2004 performed poorly in unconstrained conditions, and therefore were not applicable in real-world settings \cite{SurveyOnFD_CVIU2015}. Viola and Jones's seminal work on boosted cascaded classification-based face detection \cite{VJFull} was the first algorithm that made face detection feasible in real-world applications and is still used widely in digital cameras, smartphones and photo organization software. The method, however, works reasonably well only for near-frontal faces under normal illumination without occlusion \cite{FDSurvey_MSR}. Extensions of the boosted architecture for multi-view face detection are found in literature, such as in \cite{RotationInvMultiview_Huang}\cite{Multiview_heyden}, but these detectors are difficult to train, and do not perform well because of inaccuracies introduced by viewpoint estimation and quantization \cite{FDSurvey_MSR}. A more robust face detector is introduced in \cite{Ramanan:2012:FDP:2354409.2355119} that uses facial components or parts to construct a deformable part model (DPM) of a face. Similar geometrical modeling approaches are found in \cite{Component_Adv_Bileschi}\cite{LAEOdataset}. In \cite{Shen_ExamplarFD}, the authors introduced an examplar-based face detection method that does not require multi-scale shifting windows. As support vector machines (SVMs) became effective for classification and robust image features like SURF, local binary pattern (LFP) histogram of oriented gradient (HoG) and their variants were designed, researchers proposed different combinations of features with SVM for robust face detection \cite{SurveyOnFD_CVIU2015}. Recently, in \cite{HeadHunterMathias2014Eccv} the authors improved the performance of the DPM-based method and also introduced Headhunter, a new face detector that uses Integral Channel Features (ICF) with boosting to achieve state-of-the-art performance in face detection in the wild. A fast face detector that uses the scale invariant and bounded Normalized Pixel Difference (NPD) features is proposed in \cite{NPDDetector_2015} that uses a single soft-cascade classifier to handle unconstrained face detection. The method is claimed to achieve state-of-the-arts performance on FDDB, GENKI, and CMU-MIT datasets.

The performance break-through observed after the introduction of Deep Convolutional Neural Networks (DCNN) can be attributed to the availability of large labeled datasets, availability of GPUs, the hierarchical nature of the deep networks and regularization techniques such as dropout space \cite{SurveyOnFD_CVIU2015}. 

Continuous authentication of mobile devices requires partially visible face detection and verification to operate reliably \cite{AA02_MahbubChellappa_BTAS2016}. In \cite{FSFD_Mahbub}, the authors introduced a face detection method based on facial segments to detect partial faces on images captured for active authentication with smartphones. The idea was to produce face proposals by employing a number of weak Adaboost facial segment detectors on each image and clustering them. The authors proposed to form at most $\zeta$ subsets of facial segments from each cluster. Unique proposals were obtained by filtering out the redundant clusters that have exactly the same facial segments with the same bounding boxes. Statistical features from the unique proposals were then used to train a support vector machine classifier for face detection. The method worked well on AA-01-FD\cite{umd_Dataset} and UMDAA-02-FD\cite{AA02_MahbubChellappa_BTAS2016} mobile face detection datasets compared to other non-CNN methods \cite{AA02_MahbubChellappa_BTAS2016}. \cite{Sarkar_DeepFeatureFD} and \cite{yang2015faceness} are two other methods that explicitly address the partial face detection problem. Specially in \cite{yang2015faceness} the authors achieve state-of-the-arts performance on the FDDB, PASCAL and AFW datasets by generating face  parts responses from  an attribute-aware deep network and refining the face hypothesis for partial faces. Among other recent works, HyperFace \cite{RRanjan_Hyperface} is a deep multi-task learning framework for face detection, landmark localization, pose Estimation, and gender recognition. The method exploits the synergy among related tasks by fusing the intermediate layers of a deep CNN using a separate CNN and thereby boosting their individual performances.

\section{Proposed Method}\label{Proposed Method}

There are multiple paradigms of face detection for the mobile platform, one of which is to detect a face from facial segments. This is because of significant presence of partial faces in this domain as discussed earlier. Therefore, SegFace a simpler traditional feature-based scheme and DeepSegFace, a DCNN-based architecture, both of which detect faces from proposals composed of face segments, are proposed in this paper.

\subsection{Proposal generation}\label{section:PropGen}

\begin{figure*}[t]
\centering
\includegraphics[width=0.9\textwidth]{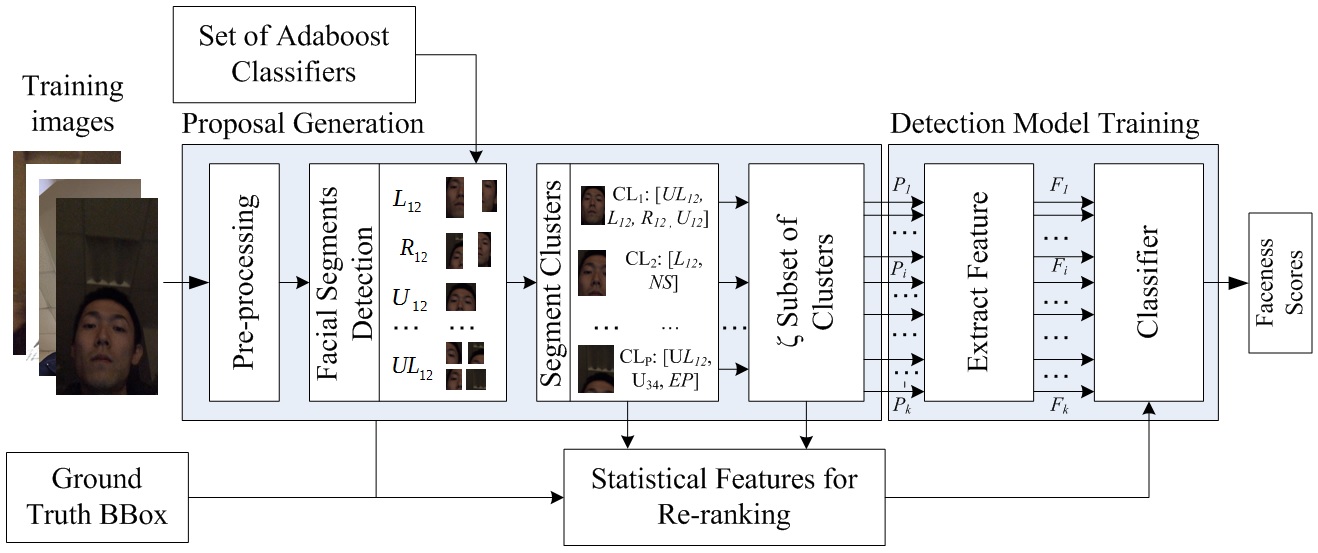}
\caption{block diagram showing the general architecture of a face segment to face detector, with components such as facial segments-based proposal generator, feature extractor, classifier and re-ranking based on prior probabilities of segments}
\label{SysteDiagramWithFeatures}
\vskip -8pt
\end{figure*}

The set of facial segments is denoted by $S = \{a_k \mid k=1,2,\dots M\}$, where $M$ is the number of segments under consideration and $a_k$ is a particular facial segment. $M$ weak Adaboost facial segment detectors are trained to detect each of the segments in $S$. After running all the segment detectors on an image, the detected face segments which produce nearby estimated face center are grouped into clusters $\text{CL}_j$, $j=\{1,2, \dots c_I\}$ as discussed in \cite{FSFD_Mahbub}. Here, $c_I$ is the number of clusters formed for image $I$. A bounding box for the whole face $B_{\text{CL}_j}$ is calculated based on the constituent segments. For the generation of a proposal set, duplicate clusters that yield exactly same bounding boxes are eliminated and at most $\zeta$ face proposals are generated from each cluster by selecting random subsets of face segments constituting that cluster.

Therefore, each proposal $P$ is composed of a set of face segments $S_P$, where $S_P \in \mathcal{P}(S)- \{\varnothing\}$ and $\mathcal{P}$ denotes the power set. To get better proposals, one can impose extra requirements such as $|S_P| > c$, where $|\cdot|$ denotes cardinality and $c$ is a threshold. Each proposal is also associated with a bounding box for the whole face, which is the smallest bounding box that encapsulates all the segments in that proposal.

In our proposal generation scheme, $M=9$ is used. The nine parts under consideration are nose ($Nose$), eye-pair ($Eye$), upper-left three-fourth ($UL_{34}$), upper-right three-fourth ($UR_{34}$), upper-half ($U_{12}$), left three-fourth ($L_{34}$), upper-left-half ($UL_{12}$), right-half ($R_{12}$) and left-half ($L_{12}$. These nine parts, constituting the best combination $C_{best}$ \cite{FSFD_Mahbub} according to the analysis of effectiveness of each part in detecting faces, are considered in this experiment since the same adaboost classifiers are adapted in this work for proposal generation. The threshold $c$ is set to $2$. A small value is chosen to get high recall, at the cost of low precision. This lets one generate a large number of proposals, so that any face is not missed in this stage. $\zeta$ is set to $10$.

The general facial segment to face detector pipeline is shown schematically in Fig. \ref{SysteDiagramWithFeatures}. The figure depicts the integration of the proposal generation block into the pipeline. In the following sections, two instances of the detection model training block shown in the figure namely, SegFace and DeepSegFace, are discussed.

\subsection{SegFace}\label{Subsec:segFace}

SegFace is a fast and shallow face detector built from segments proposal. For each segment in $s_k \in S$, a classifier $C_{s_k}$ is trained to accept features $f(s_k)$ from the segment and generate a score denoting if a face is present. Output scores of $C_{s_k}$ are stored in an $M$ dimensional feature vector $F_{C}$, where, elements in $F_{C}$ corresponding to segments that are not present in a proposal are set to $0$. 

Another feature vector of size $2M+2$ is constructed using several prior probability values as features from the training proposal set that represents the likeliness of certain segments and certain combinations. These values are

\begin{itemize}
\item Fraction of total true faces constituted by proposal $P$, i.e. 
\begin{equation}
\frac{| P \in \Theta^F |}{|\Theta^F |}, \nonumber
\end{equation} 
where $\Theta^F$ is the set of all proposals that return a true face.
\item The fraction of total mistakes constituted by proposal $P$, i.e. 
\begin{equation}
\frac{|P \in \Theta^{\overline{F}}|}{|\Theta^{\overline{F}}|}, \nonumber
\end{equation}, where, $\Theta^{\overline{F}}$ is the set of all proposals that are not faces.
\item For each of the $M$ facial segment $s_k \in S$, the fraction of total true face proposals of which $s_k$ is a part of, i.e.
\begin{equation}
\frac{|s_k \in S_p ; S_p \in \Theta^F|}{|\Theta^F|}, \text{ where, }k=1, 2, \hdots, M. \nonumber
\end{equation}
\item For each of the $M$ facial segment $s_k \in S$, the fraction of total false face proposals of which $s_k$ is a part of, i.e.
\begin{equation}
\frac{|s_k \in S_p ; S_p \in \Theta^{\overline{F}}|}{|\Theta^{\overline{F}}|},\text{ where, } k=1, 2, \hdots, M. \nonumber
\end{equation}
\end{itemize}

$F_C$ and $F_S$ are appended together to form the full feature vector $F$ of length $3M+2$. Then a master classifier $C$ is trained on the training set of such labeled vectors $\{F_i, Y_i\}$, where $Y_i$ denotes the label (face or no-face). Thus, $C$ learns how to assign relative importance to different segments and likeliness of certain combinations of segments occurring in deciding if a face is present in a proposal.
Thus, SegFace extends the face detection from segments concept in \cite{FSFD_Mahbub} using traditional methods to obtain reasonably accurate results. In our implementation of SegFace, HoG \cite{HOG_Features} features are used as $f$ and Support Vector Machine (SVM) classifiers \cite{SVM_tutorial} are used as both $C_{s_k}$ and $C$ for generating segment-wise scores, as well as the final detection score, respectively.

\subsection{DeepSegFace: CNN Architecture for detecting faces from face segments}
\begin{figure*}
\centering
\includegraphics[width=0.9\textwidth]{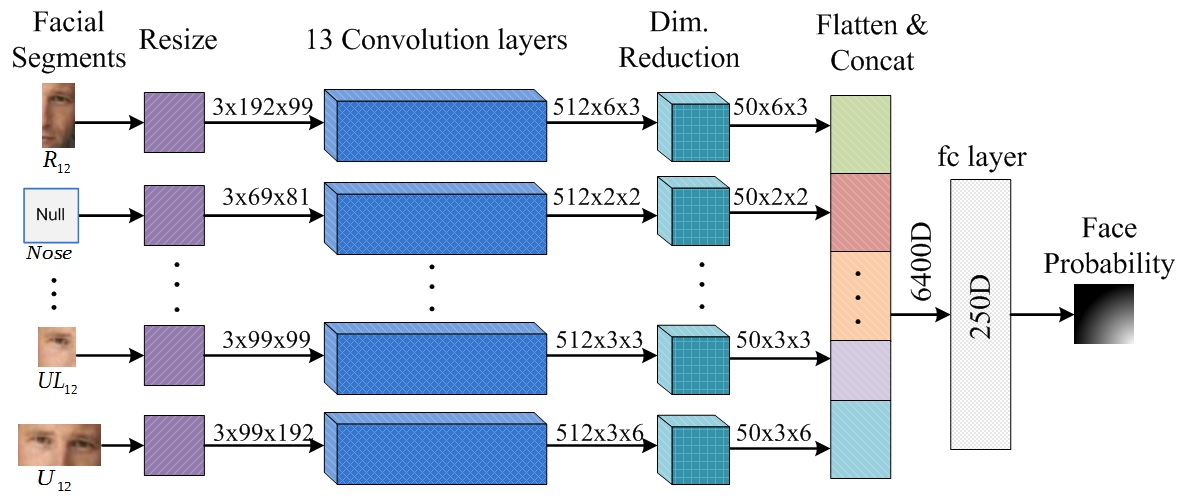}
\caption{Block diagram showing DeepSegFace architecture.}
\label{DeepSegFaceArch}
\end{figure*}

DeepSegFace is an architecture to integrate deep CNNs and segments-based face detection. Thus, it allows for end-to-end training and for exploiting the superior capabilities of deep CNNs for segment-based detection. At first, proposals, consisting of subsets of the $M=9$ parts as discussed earlier, are generated for each image. DeepSegFace is then trained to calculate the probability values of the proposal being a face. Finally, a re-ranking step adjusts the probability values from the network. The proposal with the maximum re-ranked score is deemed as the detection for that image.

The architecture of DeepSegFace is arranged according to the classic paradigm in pattern recognition: feature extraction, dimensionality reduction followed by a classifier. Training occurs end-to-end (proposal to face probability). A simple block diagram of the architecture is shown in Fig. \ref{DeepSegFaceArch}. Different components of the figure are discussed here.

\emph{Convolutional Feature Extraction}: There are nine convolutional networks for each of the nine segments. Each of the nine columns is structurally similar to and initialized with the convolution layers of VGG16 network \cite{VGG}. Thus each network has thirteen convolution layers arranged in five blocks. Each segment in the proposal is resized to standard dimensions for that segment, then the VGG mean value is subtracted from each channel. For segments not present in the proposal, zero-input is fed into the networks corresponding to those segments, as shown for the $Nose$ segment in the figure.

\emph{Dimensionality reduction}: Since the last convolutional layer of each network outputs $512$ feature maps, in total the number of features is quite large. Therefore, a randomly initialized convolutional layer with filter size $1\times1$ and fifty feature maps is added to learn an appropriate dimension reduction.

\emph{Classifier}: The outputs from the dimensionality reduction block for each segment-network are flattened and concatenated side by side to yield a $6400$ dimensional feature vector. A fully connected (fc) layer of $250$ nodes, followed by a softmax layer of two nodes (both randomly initialized) is added on top of the feature vector. The two outputs of the softmax layer corresponds to the probability of being a face or not being a face and hence they sum to one.

\emph{Re-ranking}: The DeepSegFace network outputs the face detection probabilities for each proposal in an image, which can be used to rank the proposals and then declare the highest probability proposal as the face in that image. However there is some prior knowledge that some segments are more effective at detecting the presence of faces than others. This information is available from the prior probability values discussed in section \ref{Subsec:segFace}. In the case of SegFace, the statistical features are incorporated in the feature vector of the master SVM. Since these feature are similar to fixed priors for the segments and their differnt combinations, for DeepSegFace, these values are used to re-rank the final score by multiplying it with the mean of the statistical features. 

More details on the dimensions of DeepSegFace architecture at different stages are presented in Table \ref{DeepSegFaceConvLayers}. Key insights about the structure of the network are provided in the next subsection.

\begin{table}
\centering
\caption{Structure of DeepSegFace's Convolutional layers (feature extraction and dimensionality reduction)}
\begin{tabular}{c  c c c c}
\hline
Segment         & Input &  Feature & Dim. Reduce & Flatten \\
\hline
\hline

$Nose$   & $3\times69\times81$ &  $512\times2\times2$  & $50\times2\times2$ & $200$ \\
\hline
$Eye$   & $3\times54\times162$ &  $512\times1\times5$  & $50\times1\times5$ & $250$\\
\hline
$UL_{34}$       & $3\times147\times147$ &  $512\times4\times4$  & $50\times4\times4$ & $800$\\
\hline
$UR_{34}$       & $3\times147\times 147$ &  $512\times4\times4$  &  $50\times4\times4$ & $800$\\
\hline
$U_{12}$       & $3\times99\times192$ &  $512\times3\times6$  &  $50\times3\times6$ & $900$\\
\hline
$L_{34}$       & $3\times192\times147$ &  $512\times6\times4$  & $50\times6\times4$ & $1200$\\
\hline
$UL_{12}$       & $3\times99\times99$ &  $512\times3\times3$  & $50\times3\times3$ & $450$\\
\hline
$R_{12}$       & $3\times192\times99$ &  $512\times6\times3$  & $50\times6\times3$ & $900$\\
\hline
$L_{12}$       & $3\times192\times99$ &  $512\times6\times3$  & $50\times6\times3$ & $900$\\

\hline

\hline
\end{tabular}
\label{DeepSegFaceConvLayers}
\vskip -5pt
\end{table}

\begin{table}
\centering
\caption{Comparision of the proposed methods}
\begin{tabular}{p{1.5cm} p{2.7cm} p{2.9cm}}
\hline
Component         & SegFace &  DeepSegFace   \\
\hline
\hline
Proposal Generation 	& Clustering detections from cascade classifiers for facial segments & Clustering detections from cascade classifiers for facial segments \\
\hline
Low level features       & HoG features &  Deep CNN features \\
\hline
Intermediate stage       & SVM for segment $i$ outputs a score on HoG features of segment $i$ &  Dimension reduction and concatenation to single 6400D vector \\
\hline
Final classifier       & SVM trained on scores from part SVMs and priors &  Fully connected layer, followed by a softmax layer \\
\hline
Using priors       & Used as features in the final SVM &  Used for re-ranking of face probabilities in post processing\\
\hline
Trade offs       & Fast but less accurate &  Slower but more accurate\\
\hline
\end{tabular}
\label{compareAlgos}
\vskip -5pt
\end{table}

\subsection{Interpretations}
\subsubsection{SegFace vs. DeepSegFace}
One can think of SegFace and DeepSegFace belonging to the same school of face detection algorithms, namely, detecting faces by pooling detections of facial segments. Both have broad structural similarities as discussed in table \ref{compareAlgos}. However SegFace focuses on resource constrained execution (no GPU) while DeepSegFace targets performance.

Currently both of their input proposals come from the same proposal generation scheme described in section \ref{section:PropGen}. Hence, both strategies will benefit from improved proposal generation algorithms. Here a very fast proposal generation scheme is chosen over more sophisticated ones mostly for processing speed. However if processing power is not a bottleneck, one can customize more advanced proposal generation schemes such as Faster R-CNN \cite{renNIPS15fasterrcnn} for this purpose.

\subsubsection{Facial Segment Drop-out for Better Generalization} 	
	As mentioned in the proposal generation scheme, subsets of face segments in a cluster are used to generate new proposals. For example, if a cluster of face segments contains $n$ segments and each proposal must contain atleast $c$ segments, then it is possible to generate $\sum\limits_{k=t}^n {{n}\choose{k}}$ proposals.	
Now, if all the facial segments are present, the network's task is easier. However, all the nine parts are redundant for detecting a face, since there are significant overlaps among the parts. Also, in a given face, often many segments are not detected by the weak segment detectors. Thus, one can interpret the missing segments as `dropped-out', i.e. some of the input signals are randomly missing (they are set to zero). Thus the network must be robust to face segments `dropping out' and generalize better to be able to identify faces.

\subsection{Data augmentation}
Training with subsets of detected proposals also has the additional effect of augmenting the data. It has been observed that around sixteen proposals are generated per image. Many of these proposals are actually training the network to detect the same face using different combination of segments. This is a more principled data augmentation technique compared to other commonly used methods like adding additional noise, in which case it is not explicitly clear how the augmentation is helping the network learn better. 

\section{Dataset}\label{Dataset}	
Given the sensitive nature of smartphone usage data, there has been a scarcity of large dataset of front camera images in natural settings. However, the following two datasets have been published in recent years that provided a platform to evaluate partial face detection methods in real-life scenarios.
\subsection{Active Authentication Dataset-01 (AA-01)}
The AA-01 dataset \cite{umd_Dataset} is a challenging dataset for front-camera face detection task which contains the front-facing camera face video for forty three male and seven female IPhone users under three different ambient lighting conditions: well-lit, dimly-lit, and natural daylight. In each session, the users performed five different tasks: enrollment, scrolling, picture count, document reading and picture dragging. To evaluate the face detector, face bounding boxes were annotated in a total of $8036$ frames of the fifty users. This dataset, denoted as AA-01-FD, contains $1607$ frames without faces and $6429$ frames with faces \cite{FSFD_Mahbub}, \cite{Sarkar_DeepFeatureFD}. The images in this are semi-constrained as the subjects perform a set task during the data collection period. However they are not required or encouraged to maintain a certain posture, hence the dataset is sufficiently challenging due to pose variations, occlusions and partial faces.

\subsection{University of Maryland Active Authentication Dataset-02 (UMDAA-02)}
The UMDAA-02 contains usage data of more than fifteen smartphone sensors obtained in a natural settings for an average of ten days per user \cite{AA02_MahbubChellappa_BTAS2016}. The UMDAA-02 Face Detection Dataset (UMDAA-02-FD) dataset contains a total of $33,209$ images, manually annotated for face bounding box, from all sessions of the $44$ users ($33$ male, $10$ female) of UMDAA-02 sampled at an interval of $7$ seconds. This dataset is truly unconstrained as data is collected during real-time phone usage over a period of one week. The face images have wide variation of pose and  illumination, and it is observed that the faces are mostly large in size and a good proportion of the face images are only partially visible. 

\section{Experimental Results}\label{Experimental Results}
\subsection{Experimental Setup}
The experimental results, presented in this section, demonstrate the effectiveness of the proposed methods over other state-of-the-art face detection algorithms. In particular, experimental results on the AA-01-FD and UMDAA-02-FD datasets are compared with a) Normalized Pixel Difference (NPD)-based detector \cite{NPDDetector_2015}, b) Hyperface detector \cite{RRanjan_Hyperface}, c) Deep Pyramid Deformable Part Model detector \cite{RRanjan_DeepPyramidFD}, d) DPM baseline detector \cite{HeadHunterMathias2014Eccv}, and e) Facial Segment-based Face Detector (FSFD) \cite{FSFD_Mahbub}. Both SegFace and DeepSegFace are trained on $3964$ images from AA-01-FD and trained models are validated using $1238$ images. The data augmentation process produces $57,756$ proposals from the training set, that is around $14.5$ proposals per image. The remaining $2835$ images of AA-01-FD are used for testing. For UMDAA-02-FD,  $32,642$ images are used for testing. In all experiments with SegFace and DeepSegFace, $c=2$ and $\zeta=10$ is considered.

The results are evaluated by comparing the ROC curve and precision-recall curves of these detectors since all of them return a confidence score for detection. The goal is to achieve high True Acceptance Rate (TAR) at a very low False Acceptance Rate (FAR) and also a high recall at a very high precision. Hence, numerically, the value of TAR at $1\%$ FPR and recall achieved by a detector at $99\%$ precision are the two metrics that are used to compare different methods.

\begin{table}[t]
\centering
\caption{Comparison at $50\%$ overlap on AA-01-FD and UMDAA-02-FD datasets}
\vskip -8pt
\begin{tabular}{c c c c c} 
\hline
\multirow{2}{*}{Methods} &
  \multicolumn{2}{c}{AA-01} &
  \multicolumn{2}{c}{UMDAA-02}\\
\cline{2-5}
	& TAR at 	& Recall at  & TAR at & 	Recall at	\\ 
	    & $1\%$ FAR	& $99\%$ Prec.	& $1\%$ FAR & $99\%$ Prec.	\\ 
\hline
\hline
NPD \cite{NPDDetector_2015}					& $29.51$ 	& $11.0$ 	& $33.49$ & $26.79$\\ 
\hline
DPMBaseline	\cite{HeadHunterMathias2014Eccv}			& $85.08$	& $83.25$ & $78.48$ & $72.79$	\\ 
\hline
DeepPyramid	\cite{RRanjan_DeepPyramidFD}			& $66.17$ 	& $42.35$ & $71.19$&  $66.07$	\\ 
\hline
HyperFace \cite{RRanjan_Hyperface}				& $90.52$ 	& $90.32$ &  $73.01$& $71.14$	\\ 
\hline
FSFD $C_{\text{best}}$ \cite{FSFD_Mahbub}	& $59.06$	& $55.65$ & $55.74$ & $26.88$	\\ 
\hline
SegFace					& $67.12$	& $63.09$ & $66.44$ & $61.47$	\\ 
\hline
DeepSegFace 			& $87.16$	& $86.49$ & $82.26$ &	$76.28$ \\ 
\hline
\end{tabular}
\label{AA01Dataset_Results}
\vskip -5pt
\end{table}

In table \ref{AA01Dataset_Results}, the performance of SegFace and DeepSegFace are compared with state-of-the-arts methods for both datasets. From the measures on the AA-01-FD dataset, it can be seen that SegFace, in spite of being a traditional feature based algorithm, outperforms several algorithms like FSFD and even DCNN based algorithms such as NPD and DeepPyramid. On the other hand, DeepSegFace outperforms all the other methods except HyperFace in terms of the two evaluation measures on the AA-01-FD dataset. Hyperface is a state-of-the-art algorithm that is trained on over $20,000$ images, compared to only $5202$ images used to train DeepSegFace. Also Hyperface uses R-CNN to generate face proposals, compared to the fast weak classifiers used by DeepSegFace. Furthur analysis reveals that one of the bottlenecks of DeepSegFace's performance is the proposal generation phase, thus its performance can increase if it uses a more powerful proposal generation scheme, such as R-CNN\cite{Girshick:2014:RFH:2679600.2679851}.


\begin{figure}[t]
\centering
\includegraphics[width=0.48\textwidth]{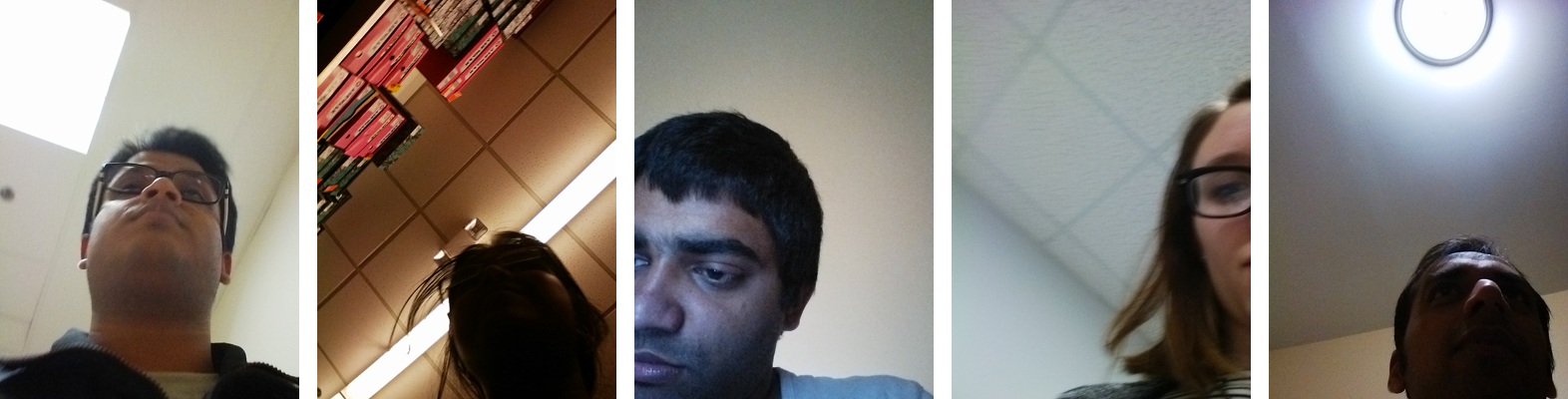}
\caption{Images without even one good proposal returned by the proposal generation mechanism. This bottleneck can be removed by using better proposal generation schemes.}
\label{imagesWithNotEvenOneGoodProposal}
\end{figure}

\begin{figure*}[t]
\centering
\includegraphics[width=0.95\textwidth]{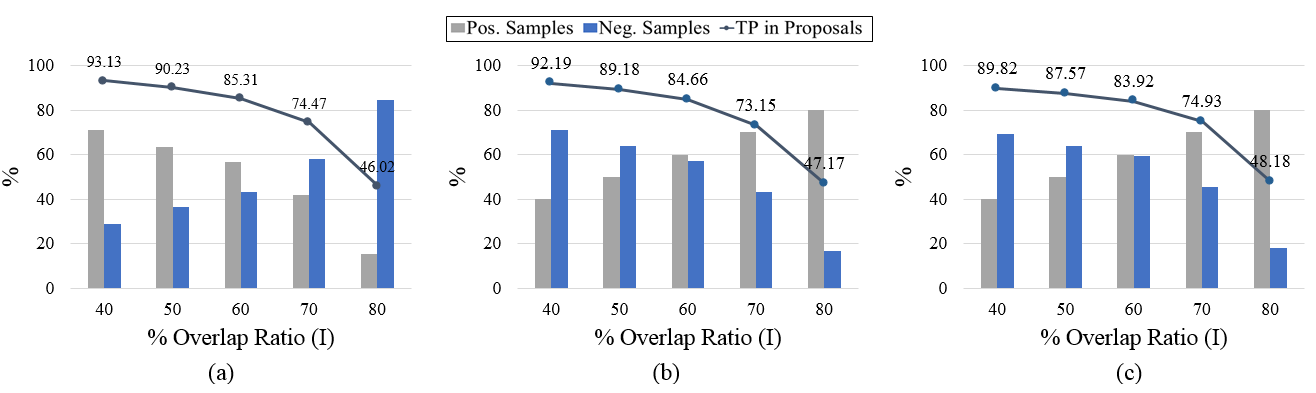}
\vskip -8pt
\caption{(a) for $57756$ Train Proposals from AA-01-FD Dataset, (b) $39168$ Test Proposals  from AA-01-FD Dataset, and (c) $410138$ Test Proposals from UMDAA-02-FD dataset. In all cases $c=2$ and $\zeta=10$.}
\label{InfoBarGraphs}
\end{figure*}

In Fig. \ref{imagesWithNotEvenOneGoodProposal}, some images are shown for which the proposal generator did not return any proposals or returned proposals without sufficient overlap, even though there are somewhat good, visible faces or facial segments in them. The percentage of true faces that are represented by at least one proposal in the list of proposals for the training and test sets are counted. The result of this analysis is shown in Fig. \ref{InfoBarGraphs}. The bar graphs denote the percentage of positive samples and negative samples present in the proposal list generated for a certain overlap ratio. For example, out of the $55,756$ proposals generated for training, there are approximately $62\%$ positive samples and $35\%$ negative samples at an overlap ratio of $50\%$. Considering the overlap ratio fixed to $50\%$ for this experiment, it can be seen from the line plot in Fig. \ref{InfoBarGraphs}(b), corresponding to the AA-01-FD test set, that the proposal generator actually represent $89.18\%$ of the true faces successfully and fails to generate a single good proposal for the rest of the images. Hence, the performance of the proposed detectors are upper-bounded by this number on this dataset, a constraint that can be mitigated by using advanced proposal schemes like R-CNN which generates around $2000$ proposals per image for Hyperface, compared to just around sixteen proposals that are generated by the fast proposal generator employed by DeepSegFace.

\begin{figure}[t]
\centering
\includegraphics[width=0.48\textwidth]{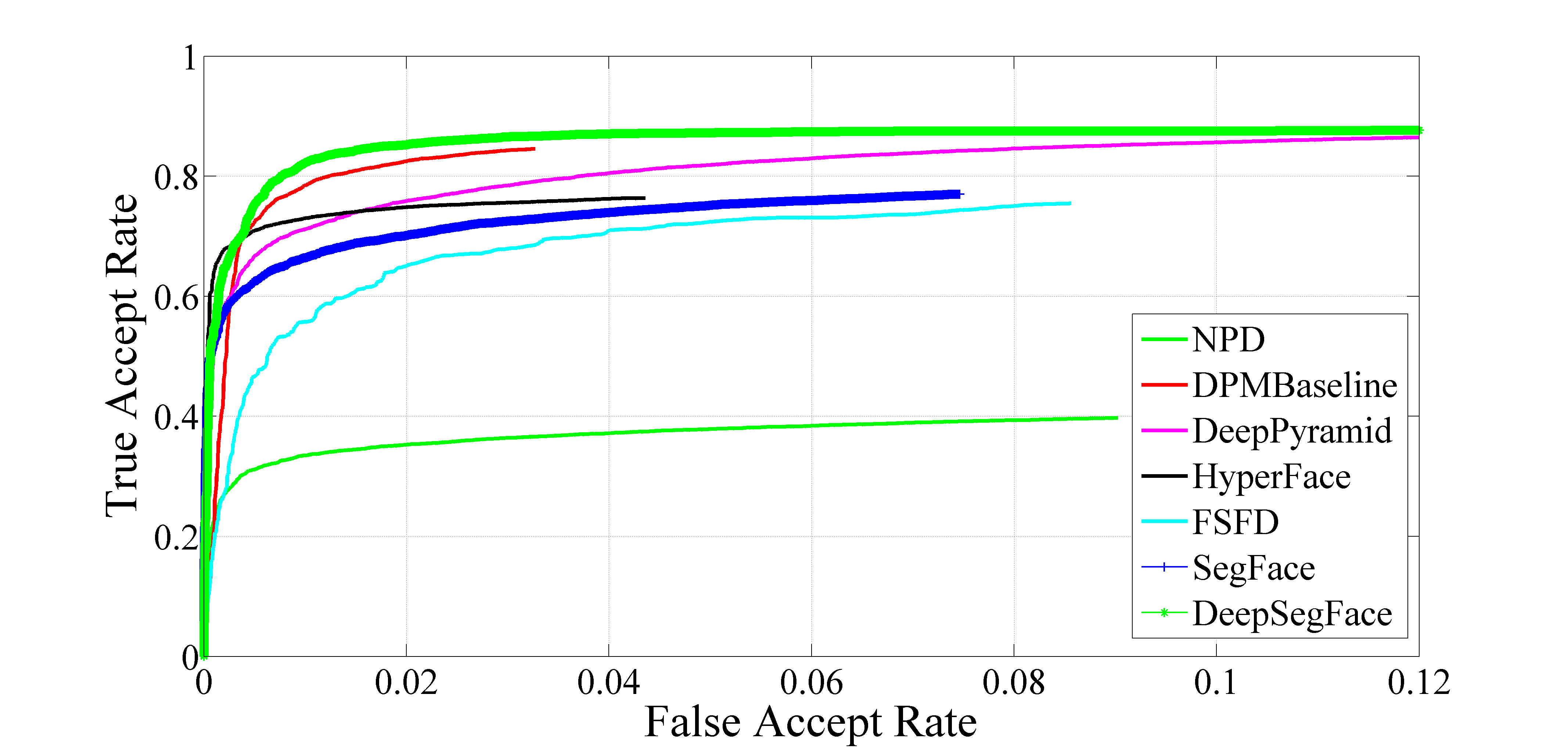}
\caption{ROC curve for comparison of different face detection methods on the UMDAA-02 dataset}
\label{ROCUMDAA02}
\end{figure}

\begin{figure}[t]
\centering
\includegraphics[width=0.48\textwidth]{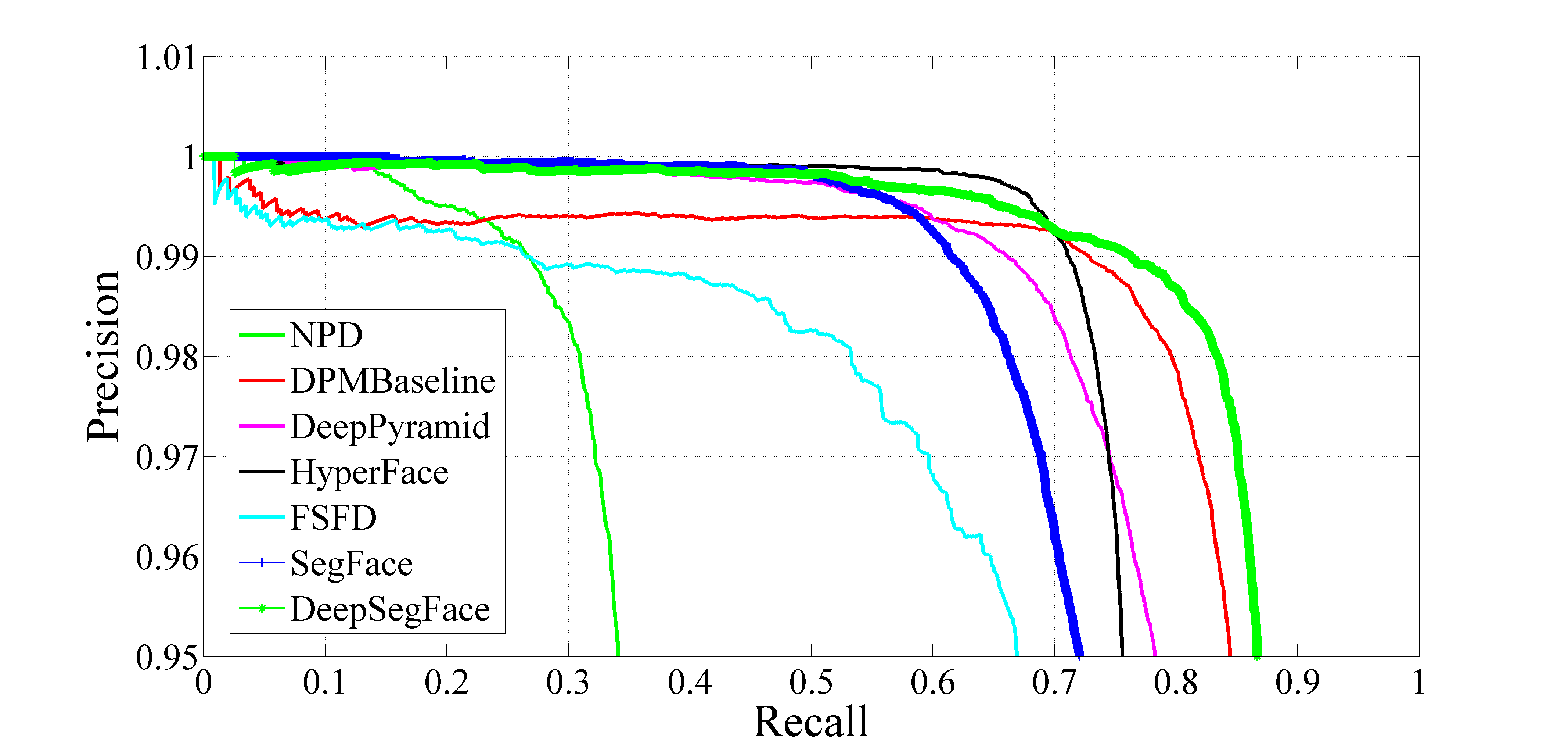}
\caption{Precision-Recall curve for comparison of different face detection methods on the UMDAA-02 dataset.}
\label{PrecisionRecallUMDAA02}
\end{figure}

However, when considering the UMDAA-02-FD test set, which is completely unconstrained and has almost ten times more images than AA-01-FD test set, this upper bound might not be so bad. From Fig. \ref{InfoBarGraphs}(c) it can be seen that the upper bound for UMDAA-02-FD is $87.57\%$ true positive value. Now, in Fig. \ref{ROCUMDAA02}, the ROC for this dataset is shown. It can be seen that the DeepSegFace method outperforms all the other methods, including HyperFace, with a large margin even with the upper bound (the curve flattens around $87.5\%$). This is because all the traditional methods suffer so much more when detecting mobile faces in truly unconstrained settings that a true acceptance rate of even $87\%$ is hard to achieve. It is to be noted that the data collection process for AA-01-FD was task-based \cite{umd_Dataset} and hence, supervised, while UMDAA-02-FD was collected in a completely natural setting \cite{AA02_MahbubChellappa_BTAS2016}. The precision-recall curve for UMDAA-02-FD dataset is shown in Fig. \ref{PrecisionRecallUMDAA02}. It can be seen that the DeepSegFace method has much better recall at $99\%$ precision than any other method. In both figures, the performance of SegFace is found satisfactory given its dependency on traditional features. In fact, the curves are not too far off from the DeepPyramid method, which is DCNN-based. 

The proposals for both test sets are analyzed to reveal that on an average only three segments per proposal are present for both datasets. Thus, while there are nine convolutional networks in the architecture, only three of them need to fire on an average for generating scores from the proposals. When forwarding proposals in batch sizes of $256$, DeepSegFace takes around $0.02$ seconds per proposal on a GTX Titan-X GPU. SegFace takes around $0.49$ seconds when running on a Intel Xeon CPU E-2623 v4 ($2.604$ GHz) machine with $32$GB Memory without multi-threading, hence it is possible to optimize it to run on mobile devices in reasonable time without specialized hardware.

\section{Conclusion and Future Directions}\label{Conclusion}
This paper proposes two schemes, DeepFaceSeg and FaceSeg, for detecting faces captured by front cameras of smartphones, primarily for the purpose of active authentication. By detecting faces from facial segments, the algorithms are well equipped to handle partial faces which are prevalent in the mobile face domain. Also a principled data augmentation for this class of algorithms is proposed that makes the network generalize well.
DeepFaceSeg, a DCNN architecture, performs very well on two mobile face datasets, while FaceSeg, which is designed for speed, works reasonably well, but is simple and fast enough to be implemented on mobile platforms for real time operations.

DeepSegFace and FaceSeg are but two instantiations of the general concept of detecting faces from proposals with facial segment. The general idea can be extended by using improved segment proposals or better classifiers. These detected faces comes with several facial segments, which opens up the research opportunity to investigate facial segment-based fiducial estimation, attribute detection and face verification. Future research  direction may also include customizing and optimizing both algorithms to implement them in mobile-platforms for real-life applications.

\section*{Acknowledgement}
This work was done in partnership with and supported by Google Advanced Technology and Projects (ATAP), a Skunk Works-inspired team chartered to deliver breakthrough innovations with end-to-end product development based on cutting edge research and a cooperative agreement FA8750-13-2-0279 from DARPA.

{\small
\bibliographystyle{ieee}
\bibliography{biblio_PFD}

\begin{thebibliography}{10}\itemsep=-1pt

\bibitem{Component_Adv_Bileschi}
S.~Bileschi and B.~Heisele.
\newblock Advances in component based face detection.
\newblock In {\em Analysis and Modeling of Faces and Gestures, 2003. AMFG 2003.
  IEEE International Workshop on}, pages 149--156, Oct 2003.

\bibitem{SVM_tutorial}
C.~J. Burges.
\newblock A tutorial on support vector machines for pattern recognition.
\newblock volume~2, pages 121--167, January 1998.

\bibitem{Bledsoe1965}
H.~Chan and W.~Bledsoe.
\newblock A man-machine facial recognition system: Some preliminary results.
\newblock 1965.

\bibitem{HOG_Features}
N.~Dalal and B.~Triggs.
\newblock Histograms of oriented gradients for human detection.
\newblock In {\em Proceedings of the 2005 IEEE Computer Society Conference on
  Computer Vision and Pattern Recognition (CVPR'05) - Volume 1 - Volume 01},
  CVPR '05, pages 886--893, Washington, DC, USA, 2005. IEEE Computer Society.

\bibitem{YahooMultiview_FD}
S.~S. Farfade, M.~J. Saberian, and L.-J. Li.
\newblock Multi-view face detection using deep convolutional neural networks.
\newblock In {\em Proceedings of the 5th ACM on International Conference on
  Multimedia Retrieval}, ICMR '15, pages 643--650, New York, NY, USA, 2015.
  ACM.

\bibitem{Girshick:2014:RFH:2679600.2679851}
R.~Girshick, J.~Donahue, T.~Darrell, and J.~Malik.
\newblock Rich feature hierarchies for accurate object detection and semantic
  segmentation.
\newblock In {\em Proceedings of the 2014 IEEE Conference on Computer Vision
  and Pattern Recognition}, CVPR '14, pages 580--587, Washington, DC, USA,
  2014. IEEE Computer Society.

\bibitem{RotationInvMultiview_Huang}
C.~Huang, H.~Ai, Y.~Li, and S.~Lao.
\newblock High-performance rotation invariant multiview face detection.
\newblock {\em Pattern Analysis and Machine Intelligence, IEEE Transactions
  on}, 29(4):671--686, April 2007.

\bibitem{LFWTech}
G.~B. Huang, M.~Ramesh, T.~Berg, and E.~Learned-Miller.
\newblock Labeled faces in the wild: A database for studying face recognition
  in unconstrained environments.
\newblock Technical Report 07-49, University of Massachusetts, Amherst, October
  2007.

\bibitem{fddbTech}
V.~Jain and E.~Learned-Miller.
\newblock Fddb: A benchmark for face detection in unconstrained settings.
\newblock Technical Report UM-CS-2010-009, University of Massachusetts,
  Amherst, 2010.

\bibitem{AFLWDataset}
M.~Kostinger, P.~Wohlhart, P.~Roth, and H.~Bischof.
\newblock Annotated facial landmarks in the wild: A large-scale, real-world
  database for facial landmark localization.
\newblock In {\em Computer Vision Workshops (ICCV Workshops), 2011 IEEE
  International Conference on}, pages 2144--2151, Nov 2011.

\bibitem{Multiview_heyden}
S.~Li, L.~Zhu, Z.~Zhang, A.~Blake, H.~Zhang, and H.~Shum.
\newblock Statistical learning of multi-view face detection.
\newblock In A.~Heyden, G.~Sparr, M.~Nielsen, and P.~Johansen, editors, {\em
  Computer Vision — ECCV 2002}, volume 2353 of {\em Lecture Notes in Computer
  Science}, pages 67--81. Springer Berlin Heidelberg, 2002.

\bibitem{NPDDetector_2015}
S.~Liao, A.~K. Jain, and S.~Z. Li.
\newblock A fast and accurate unconstrained face detector.
\newblock {\em IEEE Trans. Pattern Anal. Mach. Intell.}, 38(2):211--223, Feb.
  2016.

\bibitem{FSFD_Mahbub}
U.~Mahbub, V.~M. Patel, D.~Chandra, B.~Barbello, and R.~Chellappa.
\newblock Partial face detection for continuous authentication.
\newblock In {\em 2016 IEEE International Conference on Image Processing
  (ICIP)}, pages 2991--2995, Sept 2016.

\bibitem{AA02_MahbubChellappa_BTAS2016}
U.~Mahbub, S.~Sarkar, V.~M. Patel, and R.~Chellappa.
\newblock Active user authentication for smartphones: A challenge data set and
  benchmark results.
\newblock In {\em Biometrics Theory, Applications and Systems (BTAS), 2016 IEEE
  7th Int. Conf.}, Sep. 2016.

\bibitem{LAEOdataset}
M.~Marin-Jimenez, A.~Zisserman, M.~Eichner, and V.~Ferrari.
\newblock Detecting people looking at each other in videos.
\newblock {\em International Journal of Computer Vision}, 106(3):282--296,
  2014.

\bibitem{HeadHunterMathias2014Eccv}
M.~Mathias, R.~Benenson, M.~Pedersoli, and L.~{Van Gool}.
\newblock Face detection without bells and whistles.
\newblock In {\em ECCV}, 2014.

\bibitem{Mobio_2012}
C.~McCool, S.~Marcel, A.~Hadid, M.~Pietikainen, P.~Matejka, J.~Cernocky,
  N.~Poh, J.~Kittler, A.~Larcher, C.~Levy, D.~Matrouf, J.-F. Bonastre,
  P.~Tresadern, and T.~Cootes.
\newblock Bi-modal person recognition on a mobile phone: using mobile phone
  data.
\newblock In {\em IEEE ICME Workshop on Hot Topics in Mobile Multimedia}, July
  2012.

\bibitem{VMP_SPM_AA_2016}
V.~M. Patel, R.~Chellappa, D.~Chandra, and B.~Barbello.
\newblock Continuous user authentication on mobile devices: Recent progress and
  remaining challenges.
\newblock {\em IEEE Signal Processing Magazine}, 33(4):49--61, July 2016.

\bibitem{Ramanan:2012:FDP:2354409.2355119}
D.~Ramanan.
\newblock Face detection, pose estimation, and landmark localization in the
  wild.
\newblock In {\em Proceedings of the 2012 IEEE Conference on Computer Vision
  and Pattern Recognition (CVPR)}, CVPR '12, pages 2879--2886, Washington, DC,
  USA, 2012. IEEE Computer Society.

\bibitem{RRanjan_DeepPyramidFD}
R.~Ranjan, V.~M. Patel, and R.~Chellappa.
\newblock A deep pyramid deformable part model for face detection.
\newblock In {\em Biometrics Theory, Applications and Systems (BTAS), 2015 IEEE
  7th Int. Conf.}, pages 1--8, 2015.

\bibitem{RRanjan_Hyperface}
R.~Ranjan, V.~M. Patel, and R.~Chellappa.
\newblock Hyperface: {A} deep multi-task learning framework for face detection,
  landmark localization, pose estimation, and gender recognition.
\newblock {\em CoRR}, abs/1603.01249, 2016.

\bibitem{renNIPS15fasterrcnn}
S.~Ren, K.~He, R.~Girshick, and J.~Sun.
\newblock Faster {R-CNN}: Towards real-time object detection with region
  proposal networks.
\newblock In {\em Advances in Neural Information Processing Systems ({NIPS})},
  2015.

\bibitem{Kanade1973}
T.~Sakai, T.~Nagao, and T.~Kanade.
\newblock Computer analysis and classification of photographs of human faces.
\newblock {\em Seminar}, pages 55--62, October 1973.

\bibitem{AA_Samangouei_CNN}
P.~Samangouei and R.~Chellappa.
\newblock Convolutional neural networks for facial attribute-based active
  authentication on mobile devices.
\newblock In {\em International Conference on Biometrics Theory, Applications
  and Systems (BTAS), Arlington, VA}, Sept 2016.

\bibitem{Sarkar_DeepFeatureFD}
S.~Sarkar, V.~M. Patel, and R.~Chellappa.
\newblock Deep feature-based face detection on mobile devices.
\newblock {\em ArXiv e-prints}, abs/1602.04868, 2016.

\bibitem{Shen_ExamplarFD}
X.~Shen, Z.~Lin, J.~Brandt, and Y.~Wu.
\newblock Detecting and aligning faces by image retrieval.
\newblock In {\em Computer Vision and Pattern Recognition (CVPR), 2013 IEEE
  Conference on}, pages 3460--3467, June 2013.

\bibitem{yang2015faceness}
C.~C.~L. Shuo~Yang, Ping~Luo and X.~Tang.
\newblock From facial parts responses to face detection: A deep learning
  approach.
\newblock In {\em Proceedings of International Conference on Computer Vision
  (ICCV)}, 2015.

\bibitem{VGG}
K.~Simonyan and A.~Zisserman.
\newblock Very deep convolutional networks for large-scale image recognition.
\newblock {\em CoRR}, abs/1409.1556, 2014.

\bibitem{CUHK_FD}
Y.~Sun, X.~Wang, and X.~Tang.
\newblock Deep convolutional network cascade for facial point detection.
\newblock In {\em Proceedings of the 2013 IEEE Conference on Computer Vision
  and Pattern Recognition}, CVPR '13, pages 3476--3483, Washington, DC, USA,
  2013. IEEE Computer Society.

\bibitem{VJFull}
P.~Viola and M.~Jones.
\newblock Rapid object detection using a boosted cascade of simple features.
\newblock In {\em Computer Vision and Pattern Recognition, 2001. CVPR 2001.
  Proceedings of the 2001 IEEE Computer Society Conference on}, volume~1, pages
  I--511--I--518 vol.1, 2001.

\bibitem{SurveyOnFD_CVIU2015}
S.~Zafeiriou, C.~Zhang, and Z.~Zhang.
\newblock A survey on face detection in the wild.
\newblock {\em Comput. Vis. Image Underst.}, 138(C):1--24, Sept. 2015.

\bibitem{FDSurvey_MSR}
C.~Zhang and Z.~Zhang.
\newblock A survey of recent advances in face detection.
\newblock Technical Report MSR-TR-2010-66, Microsoft Research, June 2010.

\bibitem{umd_Dataset}
H.~Zhang, V.~Patel, M.~Fathy, and R.~Chellappa.
\newblock Touch gesture-based active user authentication using dictionaries.
\newblock In {\em Applications of Computer Vision (WACV), 2015 IEEE Winter
  Conference on}, pages 207--214, Jan 2015.

\end{thebibliography}
}

\end{document}